\pdfoutput=1
\documentclass[11pt]{article}
\usepackage{acl}
\usepackage{times}
\usepackage{latexsym}
\usepackage[T1]{fontenc}
\usepackage[utf8]{inputenc}
\usepackage{multirow}
\usepackage{microtype}
\usepackage{inconsolata}
\usepackage{graphicx}
\usepackage{arydshln}
\usepackage{booktabs}
\usepackage{enumitem}
\title{GOSU: Retrieval-Augmented Generation with Global-Level Optimized Semantic Unit-Centric Framework}
\usepackage{amssymb}
\author{
\textbf{Xuecheng Zou}\textsuperscript{$\spadesuit$}\quad
\textbf{Ke Liu}\textsuperscript{$\diamondsuit$}\quad
\textbf{Bingbing Wang}\textsuperscript{$\heartsuit$}\\
\textbf{Huafei Deng}\textsuperscript{$\spadesuit$}\quad
\textbf{Li Zhang}\textsuperscript{$\spadesuit$}\quad
\textbf{Yu Tang}\textsuperscript{$\spadesuit$}\thanks{Corresponding author: \texttt{ytang@suda.edu.cn}.}
\\[3pt]
\begin{tabular}{c}
\textsuperscript{$\spadesuit$}\,School of Future Science and Engineering, Soochow University\\
\textsuperscript{$\heartsuit$}\,School of Mathematical Sciences, Soochow University\\
\textsuperscript{$\diamondsuit$}\,School of Information Technology (Smart Campus Education Center),\\
\quad Suzhou Institute of Trade \& Commerce
\end{tabular}
\\[4pt]
\texttt{\{xczouxczou, bbwangstat1, 20245258046, 20245258048\}@stu.suda.edu.cn}\\
\texttt{ytang@suda.edu.cn}\quad \texttt{2403200509@szjm.edu.cn}\\
}
\usepackage{xurl}
\usepackage{hyperref}
\usepackage{amsmath}
\usepackage{tabularx}
\usepackage{array}
\newcolumntype{Y}{>{\centering\arraybackslash}X}

\begin{document}
\maketitle
\captionsetup[figure]{name={Fig.},labelsep=period} 
\begin{abstract}
  Building upon the standard graph-based Retrieval-Augmented Generation (RAG), the introduction of heterogeneous graphs and hypergraphs aims to enrich retrieval and generation by leveraging the relationships between multiple entities through the concept of semantic units (SUs). But this also raises a key issue: The extraction of high-level SUs limited to local text chunks is prone to ambiguity, complex coupling, and increased retrieval overhead due to the lack of global knowledge or the neglect of fine-grained relationships. To address these issues, we propose GOSU, a semantic unit-centric RAG framework that efficiently performs global disambiguation and utilizes SUs to capture interconnections between different nodes across the global context. In the graph construction phase, GOSU performs global merging on the pre-extracted SUs from local text chunks and guides entity and relationship extraction, reducing the difficulty of coreference resolution while uncovering global semantic objects across text chunks. In the retrieval and generation phase, we introduce hierarchical keyword extraction and semantic unit completion. The former uncovers the fine-grained binary relationships overlooked by the latter, while the latter compensates for the coarse-grained $n$-ary relationships missing from the former. Evaluation across multiple tasks demonstrates that GOSU outperforms the baseline RAG methods in terms of generation quality. Our code is
available at \url{https://github.com/xczouxczou/GOSU}.
\end{abstract}

\section{Introduction}

\indent With the explosion of data scale~\citep{ouyang2022instructgpt}, the performance of large language model (LLM) is improving by leaps and bounds~\citep{achiam2023gpt4, touvron2023llama2, mei2025contextengineering},  yet their finite parameters still lead to frequent hallucinations~\citep{mallen2022whennottotrust, min2023factscore, ji2022hallucination, huang2023llmhallucination}. To this end, Retrieval-Augmented Generation (RAG)~\citep{lewis2020rag, gao2023ragsurvey, fan2024ragmeetingllms, hu2025debatersrag, asai2023selfrag}, which integrates external knowledge sources to enhance factual consistency and generation accuracy~\citep{sudhi2024ragex, es2024ragas, salemi2024evaluating, zhao2023surveyllm, tu2024reval, tonmoy2024hallucinationmitigation, shrestha2024fairrag, liu2023lostinthemiddle}, has emerged as a promising solution. In standard RAG methods, the simple approach of processing fixed-length text chunks often fails to effectively capture direct or indirect relationships between entities, limiting its practicality in knowledge-intensive tasks~\citep{pan2023llmkgroadmap, luo2023reasoningongraphs, wang2024kgpromptingmdqa, han2024graphrag, wen2023mindmap}. 

\begin{figure}[t]
  \includegraphics[width=\columnwidth]{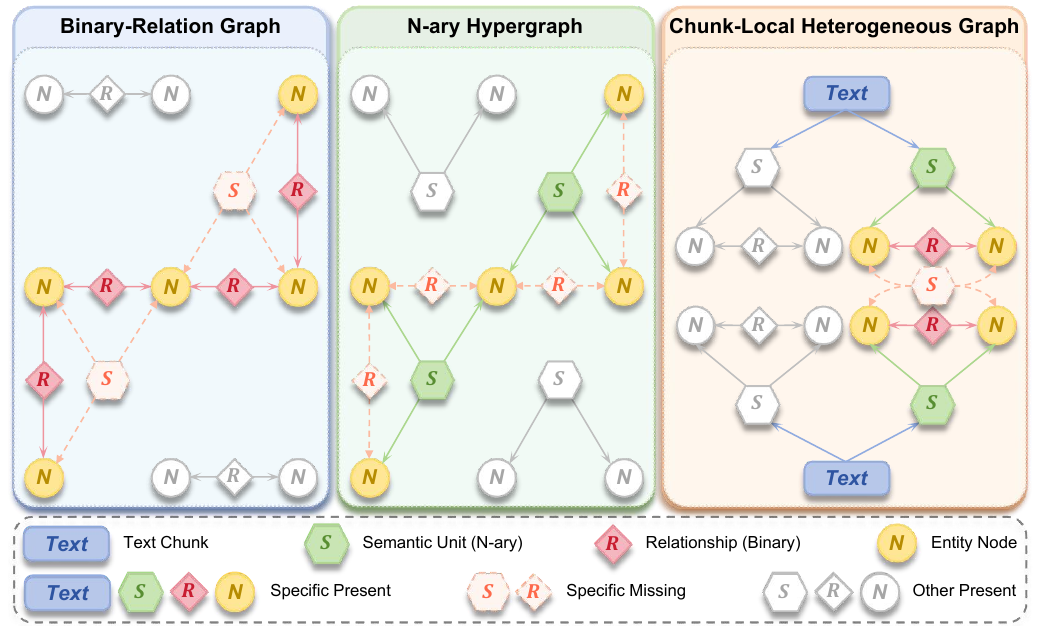}
  \caption{Comparison of graph‐based RAG structures. Binary‐Relation Graphs capture only pairwise links between entity nodes, omitting higher‐order semantic units. $N$‐ary Hypergraphs model multi‐entity events via semantic units but lack fine‐grained binary relations. Chunk‐Local Heterogeneous Graphs integrate both binary and $n$‐ary connections within isolated text chunks, yet fail to align semantic units globally across the corpus.}
  \label{fig:1}
\end{figure}

\indent Recently, graph-structured RAG methods have enhanced the ability of relational representation by incorporating knowledge graphs~\citep{edge2024graphragqfs, zhang2025rakg, liang2025kag, guo2024lightrag, tian2024gnpllm, park2023graphguided, jimenezgutierrez2024hipporag, he2024gretriever, trajanoska2023kgcllm, sanmartin2024kgrag, wang2024kgpromptingmdqa, rampasek2022gps}, but they are constrained by the binary relations inherent in structuring natural language into graphs, preventing them from effectively modeling $n$-ary relations among multiple entities and thus limiting their performance on complex reasoning tasks~\citep{wen2016beyondbinary}. Current studies are exploring the introduction of heterogeneous graphs and hypergraphs to tackle this issue~\citep{xu2025noderag, luo2025hypergraphrag, huang2025hierarchicalrag, wang2025pikerag, ma2025llmkgqa, mei2025contextengineering}. However, as illustrated in Fig.~\ref{fig:1}, decomposing events within isolated text chunks~\citep{xu2025noderag} and over-emphasizing $n$-ary relations~\citep{luo2025hypergraphrag} not only leads to information fragmentation and contextual discontinuity, but also neglects the precise representation of fine-grained relations and increases the complexity of information coupling. In other words, meeting this challenge requires optimizing the entire RAG pipeline—from knowledge graph construction through retrieval and generation—by integrating global context while balancing both coarse-grained and fine-grained relations.

\indent To address these shortcomings, we propose the GOSU framework, a RAG approach that refines semantic unit extraction at the global level and drives the entire pipeline around these semantic units (SUs). GOSU optimizes SUs at the global level through a multi-round semantic unit global merging strategy to prevent the relation fragmentation that can arise from relying on individual text chunks. Specifically, we leverage the LLM's advanced natural language processing capabilities to  identify SUs for each text block, so as to avoid the loss of critical semantic information. These identified SUs serve as pre-SUs, laying the foundation for subsequent disambiguation and deduplication merging, and ensuring semantic consistency across different text blocks. Unlike traditional graph methods based on binary relations or hyperedges, GOSU focuses on SUs—using semantic unit-centric connections during knowledge graph construction and retrieval to uncover coarse-grained $n$-ary relationships while preserving fine-grained binary relations among low-level entities, thus avoiding excessive information coupling. 

\indent Our contributions can be summarized as follows:

\begin{itemize}
  \item \textbf{Global-Level Semantic Unit Optimization: }A semantic unit global merging strategy that leverages LLM is  proposed to extract SUs from each text block and then performs global disambiguation, deduplication, and merging to ensure semantic consistency across chunks and avoid relationship fragmentation caused by local segmentation.

  \item \textbf{Semantic Unit-Centric Knowledge Graph Construction: }Diverging from traditional binary-relation or hyperedge approaches, we center the graph around SUs. This allows us to simultaneously capture coarse-grained $n$-ary relations and preserve fine-grained binary relations among underlying entities, achieving a balanced representation that mitigates over-coupling of information.

  \item \textbf{Dual-Phase Retrieval-Augmented Generation Framework: }Hierarchical keyword extraction with SU completion in both retrieval and generation stages are integrated—where keyword extraction targets fine-grained entity/term retrieval and SU completion fills in coarse-grained multi-entity SUs. The synergistic fusion of these components significantly enhances contextual coverage and generation fidelity.
\end{itemize}

\indent Experiments across multiple open knowledge intensive fields demonstrate that GOSU has superior performance in authenticity, comprehensiveness, diversity and empowerment~\citep{guo2024lightrag, qian2024memorag}, which validates that our framework provides an innovative idea for the global-level semantic unit–centric graph construction and retrieval generation paradigm, and highlights its promising potential for real-world applications.

\section{Related work}

\subsection{Retrieval-Augmented Generation}

\indent Retrieval-Augmented Generation (RAG) grounds large language model (LLM) outputs in external corpora retrieval, which aligns the generation with trusted knowledge and reducing hallucinations~\citep{huang2023llmhallucination, niu2024ragtruth, bai2024mllmhallucination, bang2025hallulens} in knowledge-intensive tasks~\citep{gao2023precise}. Subsequent improvements introduced joint training of passage-generation~\citep{izacard2022ralm} and multidimensional adaptive trade-offs~\citep{min2021jointpassage} to refine both query formulation and result filtering. Self-RAG and other variants~\citep{trivedi2022ircot, shao2023iterativesynergy, asai2023selfrag, yu2024autorag} leverage the LLM itself within iterative retrieval–reasoning loops to generate follow-up queries and assess retrieved evidence, whereas DRAG~\citep{hu2025debatersrag} and FLARE~\citep{jiang2023active, su2024dragin} introduce multi-agent debates and token-triggered retrieval to further curb hallucination and reduce unnecessary context. Despite these advances, all of these "flat" RAG approaches rely on coarse chunking and simple retrieval strategies, which tend to fragment context and inject noise when handling nuanced multi-entity events.

\subsection{Graph-Structured RAG}

\indent To better preserve relationships between entities, graph-structured RAG methods integrate knowledge graphs and graph algorithms into retrieval and generation~\citep{kim2023toc, peng2024graphragsurvey, xiang2025whentousegraphs, zhang2025graphragcustom}. Works such as GraphRAGG~\citep{edge2024graphragqfs, jiang2024ragraph, mavromatis2024gnnrag, he2024gretriever, zhang2025rakg} propagate contextual signals across retrieved text via multi-round message passing, extract entities and build binary relationship diagrams. Subsequently, KAG~\citep{liang2025kag} and LightRAG~\citep{guo2024lightrag} introduced confidence-based edge weighting mechanisms to prioritize and reinforce key relationships. GNPLLM and others~\citep{tian2024gnpllm, shen2024gear, barmettler2025conceptformer, xu2025aligngrag, luo2025gfmrarg} further fuse the learned graph embeddings with LLM representations, enriching the input features of downstream generation models. These methods have achieved significant accuracy improvements on fact-alignment tasks in dealing with paired entity relationships. However, because they are limited to modeling only binary edges, they inherently struggle to capture $n$-ary relations that span three or more entities, resulting in the loss of key information in complex event scenarios involving multi-party interactions.

\subsection{Heterogeneous and Hypergraph-Based Approaches}

\indent Moving beyond binary graphs, recent work has explored heterogeneous graph structures and hypergraphs to encode $n$-ary relations. NodeRAG~\citep{xu2025noderag} represents events as isolated nodes enriched with detailed type information, allowing the model to distinguish among entities, semantic units, and high-level summaries. HyperGraphRAG~\citep{luo2025hypergraphrag} further generalizes this approach by introducing hyperedges that directly connect multiple entities within a single relational fact, thereby preserving the integrity of $n$-ary events. HierarchicalRAG and similar systems~\citep{chen2024kgretriever, huang2025hierarchicalrag, jiao2025hirag, wang2025archrag, zou2025gtr} organize knowledge and enable hierarchical retrieval through multi-level or coarse-to-fine graph structures, while PikeRAG and others~\citep{sun2019pullnet, asai2019wikipediaroutes, ma2022openqaheterogeneous, wang2025pikerag, wei2024instructrag, six2025decompositional} align retrieval and generation more closely around the reasoning chain to highlight prominent multi-entity patterns in the graph. Although these methods substantially increase expressivity and multi-hop reasoning capabilities, they often fragment events across disconnected graph components, inflating the size and density of the index. This in turn raises retrieval overhead and can lead to incoherent generation when the model struggles to traverse highly entangled structures.

\subsection{Semantic Unit Extraction and Global Context Modeling}

Some prior efforts seek to preserve event coherence via chunk-level or sentence-level segmentation~\citep{xu2025noderag, michelmann2025segmentnarrative, yu2023longdocsegmentation, brunato2023coherent, zhao2024longragdual, liu2025passagesegmentation, ni2025crossformer} or overlapping sliding-window retrieval~\citep{lewis2020rag, izacard2020genreader, karpukhin2020dpr, wang2024bestpracticesrag}, but these remain fundamentally local strategies that lack corpus-wide consistency. No existing framework simultaneously (1) extracts semantically coherent units at a global level, (2) disambiguates and merges overlapping units across chunks, and (3) constructs a unified graph that balances coarse-grained $n$-ary relations with fine-grained binary links. Our GOSU framework fills this gap: it first leverages LLMs to identify and globally filter semantic units (SUs) across all text blocks, then builds an SU-centric graph that preserves both detailed entity links and richer multi-entity structures, and finally applies a dual-phase RAG pipeline—fine-grained keyword retrieval followed by SU completion—to achieve enhanced factuality, coherence, and coverage. 

\begin{figure*}
  \centering
  \includegraphics[width=1\textwidth]{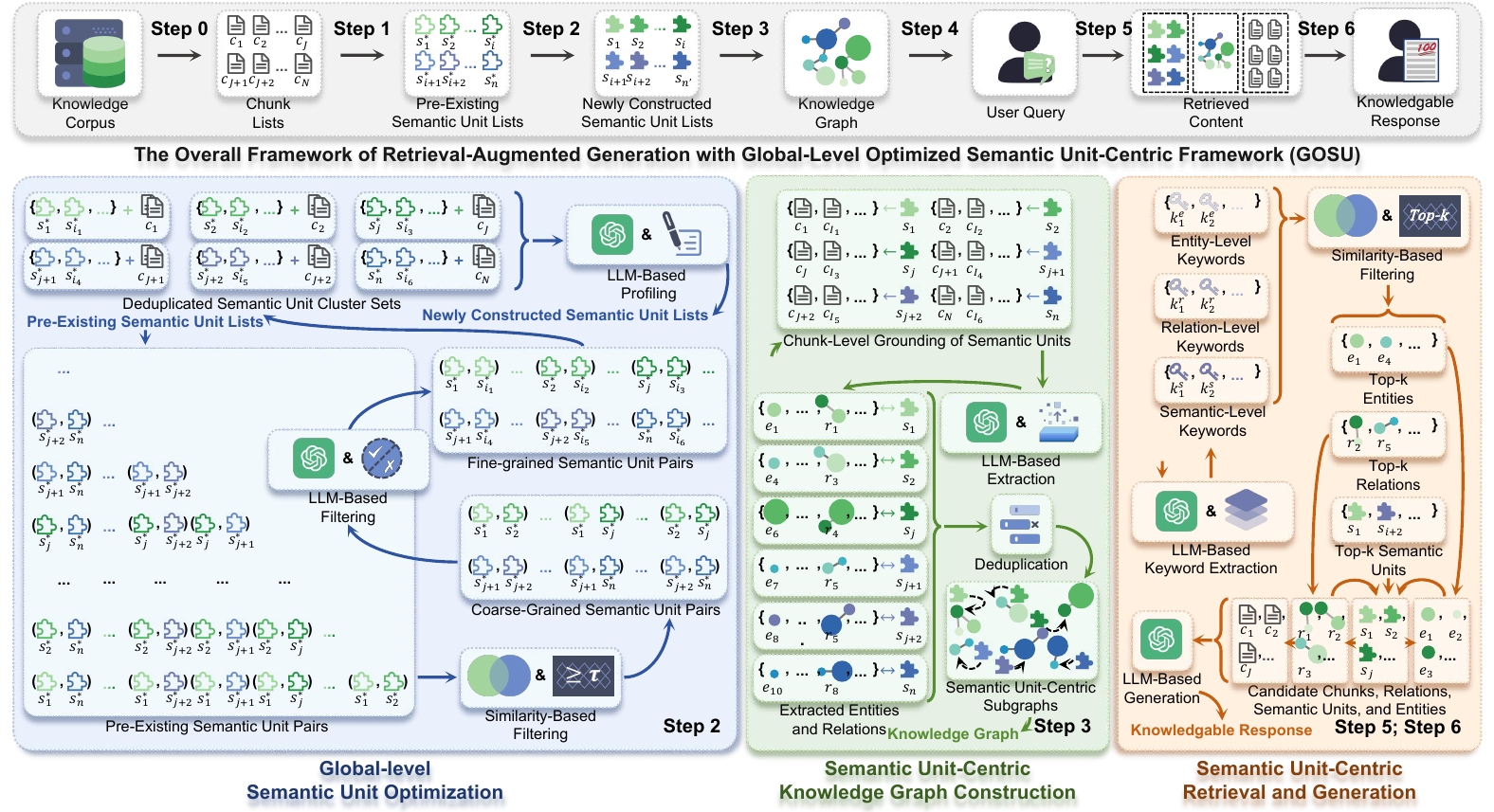}
  \caption{An overview of GOSU, which extracts semantic units from the corpus and applies global-level optimization to guide binary-relation extraction during knowledge construction, performs three-layer retrieval based on user queries at retrieval-and-generation time, and generates responses that balance fine-grained binary facts and coarse-grained $n$-ary relations..
}
  \label{fig:2}
\end{figure*}

\section{Methodology}

\indent In this section, we present global-level optimized semantic unit-centric RAG (GOSU), as illustrated in Figure~\ref{fig:2}, which comprises three core components: global semantic unit optimization, semantic unit-centric knowledge graph construction, and semantic unit-centric retrieval and generation. Detailed descriptions of each component follow in Subsections ~\ref{sec:Global Semantic Unit Optimization}, ~\ref{sec:Semantic Unit-Centric Knowledge Graph Construction} and ~\ref{sec:Semantic Unit-Centric Retrieval and Generation}, respectively.

\subsection{Global-level Semantic Unit Optimization}\label{sec:Global Semantic Unit Optimization}

\indent To ensure consistency and completeness of semantic units at the global level, GOSU introduces a “Global Semantic Unit Optimization” module at the very beginning of the pipeline, comprising three steps: initial extraction, global filtering and disambiguation, and merging with deduplication.

\paragraph{Initial Extraction} The external corpus consists of multiple documents. Each document $\mathcal{D}$ is segmented into length-controlled, semantically coherent text chunks $\mathcal{C}=\{c_{1},c_{2},\dots,c_{N}\}$ via a sliding‐window algorithm $\mathrm{DocSplit}(\cdot)$ and convert each chunk into its vector representation:

\begin{equation}
\label{eq:1}
  \mathcal{C} = \mathrm{DocSplit}(\mathcal{D}),
\end{equation}

\begin{equation}
\label{eq:2}
  \text{s.t.}\forall i \neq j, c_{i} \cap c_{j} \neq \emptyset \land \bigcup_{i=1}^{N}c_{i}=\mathcal{D}.
\end{equation}

\indent For each text chunk $c_i$ $(i=1,\ldots,N)$, we employ the selected LLM to extract a set of candidate semantic units:

\begin{equation}
\label{eq:3}
  \mathcal{S}_{i}^{*} = \{s_{i,1}^{*},s_{i,2}^{*},\dots,\,s_{i,k_i}^{*}\} \sim \mathrm{SemExt}_{\text{LLM}}(c_{i}),
\end{equation}

\noindent where $s_{i,j}^{*}$ represents the $j$-th candidate unit in chunk $i$ (an event or concept that satisfies
completeness, coherence, and information-bearing capacity), 
$k_i \!=\! |\mathcal{S}_i^{*}|$ is the (chunk-dependent) number of candidates returned for $c_i$, 
and $\mathrm{SemExt}_{\text{LLM}}(\cdot)$ is the LLM-based extraction procedure. Next, we merge the candidates from all chunks into a global pool:

\begin{equation}
\label{eq:4}
  \mathcal{S}^{*} = \bigcup_{i=1}^{N}\mathcal{S}_{i}^{*}.
\end{equation}

\noindent
where $N$ is the total number of chunks and $\mathcal{S}^{*}$ is the global candidate set.
\paragraph{Global Filtering}We first conduct a coarse filtering step based on cosine similarity. 
Given a similarity threshold, we form candidate semantic-unit pairs and then refine them with the LLM to obtain fine-grained filtering:

\begin{equation}
\label{eq:5}
\begin{aligned}
\mathrm{Coarse}(\mathcal{S}^*) 
&= \bigl\{(s_i^*,s_j^*) \;\big|\; 
   i \neq j,\;
   s_i^*,s_j^* \in \mathcal{S}^*,\\
&\qquad\;
   \mathrm{SimJudge}(s_i^*,s_j^*;\tau) = 1
\bigr\},
\end{aligned}
\end{equation}

\begin{equation}
\label{eq:6}
\begin{aligned}
\mathrm{Fine}(\mathcal{S}^*) 
&= \bigl\{(s_i^*,s_j^*) \;\big|\; 
   (s_i^*,s_j^*) \in \mathrm{Coarse}(\mathcal{S}^*),\\
&\qquad\;
   \mathrm{LLMJudge}(s_i^*,s_j^*;\tau) = 1
\bigr\},
\end{aligned}
\end{equation}

\noindent
where $\mathcal{S}^*$ is the global candidate set of semantic units,  
$(s_i^*,s_j^*)$ is an ordered pair of distinct units,  
$\mathrm{SimJudge}(\cdot,\cdot;\tau)$ is a cosine-similarity-based binary decision function with threshold $\tau$,  
and $\mathrm{LLMJudge}(\cdot,\cdot;\tau)$ denotes the LLM-based evaluator applied to pairs surviving the coarse stage.

Finally, we cluster and deduplicate the fine-grained pairs to obtain the refined semantic-unit set:

\begin{equation}
\label{eq:7}
  \hat{\mathcal{S}} = 
  \mathrm{Deduplicate}\!\circ\!\mathrm{Cluster}\!\bigl(\mathrm{Fine}(\mathcal{S}^{*})\bigr).
\end{equation}

\noindent
where $\mathrm{Cluster}(\cdot)$ groups highly similar units into clusters and  
$\mathrm{Deduplicate}(\cdot)$ removes redundant elements within or across clusters.

\paragraph{Disambiguation and Merging with Deduplication}  
In addition to retrieving the corresponding text chunks via their identifiers, 
we augment the retrieval process with vector-similarity search, thereby providing 
the LLM with sufficient evidence to interpret and integrate the semantic units.

\begin{equation}
\label{eq:8}
\begin{aligned}
  \mathrm{Retriever}_{\text{id}}(\hat{s}_{i})
  &= \bigl\{\,c_{j}\in\mathcal{C}\;\big|\; \\
  &\quad ID(c_{j}) \cap ChunkID(\hat{s}_{i}) \neq \emptyset
  \bigr\},
\end{aligned}
\end{equation}

\begin{equation}
\label{eq:9}
\begin{aligned}
  \mathrm{Retriever}_{\text{sim}}(\hat{s}_{i})
  &= \bigl\{\,c_{j}\in\mathcal{C}\;\big|\; \\
  &\quad c_{j}\notin \mathrm{Retriever}_{\text{id}}(\hat{s}_{i}), \\
  &\quad c_{j}\in \mathrm{SimRank}_{\mathcal{C}}(\hat{s}_{i})
  \bigr\}.
\end{aligned}
\end{equation}

\noindent
Here $\mathcal{C}$ denotes the full chunk set, 
$\hat{s}_{i}$ is a refined semantic unit, 
$ID(c_j)$ gives the chunk identifier of $c_j$, 
$ChunkID(\hat{s}_{i})$ is the set of chunk IDs associated with $\hat{s}_{i}$,
and $\mathrm{SimRank}_{\mathcal{C}}(\hat{s}_{i})$ returns the similarity-ranked 
neighbors of $\hat{s}_{i}$ within $\mathcal{C}$.

To avoid excessive retrieval, we prioritize ID-based lookup and then supplement it with similarity-based retrieval. The combined set is trimmed to a bounded size:

\begin{equation}
\label{eq:10}
\begin{split}
  &\mathrm{Trim}\!\left(
     \mathrm{Retriever}_{\text{id}}(\hat{s}_{i}) \cup
     \mathrm{Retriever}_{\text{sim}}(\hat{s}_{i})
   \right) \;\le \\[2pt]
  &\qquad \max\!\bigl(\tau,\,
    \mathrm{Len}(\mathrm{Retriever}_{\text{id}}(\hat{s}_{i}))
  \bigr),
\end{split}
\end{equation}

\noindent
where $\mathrm{Trim}(\cdot)$ restricts the number of retrieved chunks, 
$\tau$ is a predefined retrieval threshold, and 
$\mathrm{Len}(\cdot)$ returns the cardinality of a set.

\indent Finally, the retrieved text is integrated, and the LLM is used to globally refine the semantic units:

\begin{equation}
\label{eq:11}
\begin{aligned}
\mathcal{S} &= \mathrm{GloRef}_{\text{LLM}}\Bigl(
  \hat{\mathcal{S}},\, \mathrm{Retriever}_{\text{id}}(\hat{s}_{i}),\\
&\qquad\quad \mathrm{Retriever}_{\text{sim}}(\hat{s}_{i})
\Bigr),
\end{aligned}
\end{equation}
\noindent
where $\hat{\mathcal{S}}$ is the set of deduplicated semantic units and 
$\mathrm{GloRef}_{\text{LLM}}(\cdot)$ denotes the LLM-based global refinement procedure.

\subsection{Semantic Unit-Centric Knowledge Graph Construction}\label{sec:Semantic Unit-Centric Knowledge Graph Construction}

\indent After completing the global semantic-unit optimization, GOSU uses the refined set $\mathcal{S}$ to construct the knowledge graph via three stages: entity–relation extraction, subgraph construction, and graph assembly.

\paragraph{Entity–Relation Extraction}
Each global semantic unit $s_i\in\mathcal{S}$ is mapped to a graph node. 
For every $s_i$, an LLM extracts fine-grained entities $\mathcal{E}_i$ and binary relations $\mathcal{R}_i$, which are used to create entity nodes and relation edges while preserving context indices. 
Before assembly, we also extract locally identifiable entities and relations from each chunk to form a preliminary subgraph:

\begin{equation}
\label{eq:12}
  Pre\text{-}KG = \bigcup_{c_i\in \mathcal{C}} \mathrm{EntRelExt}(c_i),
\end{equation}

\noindent
where $\mathcal{C}$ is the set of all text chunks and $\mathrm{EntRelExt}(c_i)$ returns a set of entity–relation assertions extracted from chunk $c_i$.

For each semantic unit, entities and relations are further gathered from its supporting chunks:
\begin{equation}
\label{eq:13}
  \mathcal{E}_{i} = 
  \bigcup_{\mathrm{ID}(c_{j})\in \mathrm{ChunkID}(s_{i})}
  \mathrm{EntExt}_{\text{sem}}(s_{i},\,c_{j}),
\end{equation}

\begin{equation}
\label{eq:14}
  \mathcal{R}_{i} = 
  \bigcup_{\mathrm{ID}(c_{j})\in \mathrm{ChunkID}(s_{i})}
  \mathrm{RelExt}_{\text{sem}}(s_{i},\,c_{j}),
\end{equation}

\noindent
where $s_i$ is a semantic unit, $\mathrm{ID}(c_j)$ gives the identifier of chunk $c_j$, and $\mathrm{ChunkID}(s_i)$ is the set of chunk IDs associated with $s_i$; 
$\mathrm{EntExt}_{\text{sem}}(s_i,c_j)$ and $\mathrm{RelExt}_{\text{sem}}(s_i,c_j)$ denote the LLM-based, $s_i$-conditioned entity and relation extractors applied to the context of $c_j$, respectively.

\paragraph{Subgraph Construction}
For each global semantic unit $s_i$, we first resolve ambiguity and remove duplicates among its associated entities $\mathcal{E}_i$ and relations $\mathcal{R}_i$. 
We then build an entity–relation subgraph centered at $s_i$, preserving both binary and higher-order ($n$-ary) structure:

\begin{equation}
\label{eq:15}
  \mathcal{G}_{i}^{*} =
  \mathrm{Deduplicate}\circ \mathrm{Link}_{\text{sub}}(s_{i},\,\mathcal{E}_{i},\,\mathcal{R}_{i}),
\end{equation}

\noindent
where $\mathrm{Link}_{\text{sub}}(s_i,\mathcal{E}_i,\mathcal{R}_i)$ builds an $s_i$-centric subgraph
by linking $s_i$ to entities in $\mathcal{E}_i$ and instantiating relations in $\mathcal{R}_i$
(binary edges; $n$-ary, if any, via a small relation node).
$\mathrm{Deduplicate}(\cdot)$ merges co-referent entities/relations and removes duplicates.
$\mathcal{G}_i^{*}$ denotes the resulting cleaned subgraph.

\paragraph{Knowledge Graph Assembly}
Once all semantic-unit–centric subgraphs $\mathcal{G}_i^{*}$ are constructed, we assemble them into the final knowledge graph $\mathcal{G}$ by resolving cross-subgraph ambiguities and removing duplicates.

\begin{equation}
\label{eq:16}
  \mathcal{G} = \mathrm{Deduplicate}\circ \mathrm{Link}_{\text{all}}(\mathcal{G}^{*}),
\end{equation}

\noindent
where $\mathcal{G}^{*}=\{\mathcal{G}_i^{*}\}$ denotes the set of all subgraphs; 
$\mathrm{Link}_{\text{all}}(\mathcal{G}^{*})$ establishes cross-subgraph links by aligning co-referent entities/relations and adding inter-unit edges based on shared identifiers and context; 
$\mathrm{Deduplicate}(\cdot)$ then collapses remaining duplicates and resolves conflicts to produce $\mathcal{G}$.

\subsection{Semantic Unit-Centric Retrieval and Generation}
\label{sec:Semantic Unit-Centric Retrieval and Generation}

\indent Once the knowledge graph is built, GOSU adopts two complementary retrieval–generation pathways to jointly capture fine-grained binary relations and global $n$-ary events.

\paragraph{Hierarchical Keyword Extraction}
Building on LightRAG~\cite{guo2024lightrag}, we perform hierarchical keyword extraction from the user query $q$ to support low-cost, effective retrieval of fine-grained relations. 
Beyond prior work that uses only low-level entity keywords $\mathcal{K}_{low}$ and high-level thematic keywords $\mathcal{K}_{high}$, we introduce a mid-level “semantic-unit” tier $\mathcal{K}_{sem}$, whose compact phrases encapsulate self-contained facts, relations, or events and thus improve retrieval precision with negligible overhead:

\begin{equation}
\label{eq:17}
\begin{aligned}
(\mathcal{K}_{low},\, \mathcal{K}_{sem},\, \mathcal{K}_{high})
&= \mathrm{KeyExt}_{\text{LLM}}\!\bigl(q;\,\mathcal{G}\bigr),
\end{aligned}
\end{equation}

\noindent
where $q$ is the input query, $\mathcal{G}$ is the constructed knowledge graph (used for optional conditioning and normalization), 
$\mathcal{K}_{low}$ collects entity-/attribute-level terms (e.g., names, IDs, types), 
$\mathcal{K}_{sem}$ collects short semantic-unit phrases that summarize atomic facts or events, 
and $\mathcal{K}_{high}$ collects theme-/topic-level terms. 
Each $\mathcal{K}_{\bullet}$ is a (ranked) set of keywords yielded by the LLM extractor $\mathrm{KeyExt}_{\text{LLM}}(\cdot)$.

\paragraph{Semantic‐Unit Completion}
We first use low- and high-level keywords to locate target entities and relations, 
then enrich them with weakly related but semantically relevant nodes, edges, and chunks.
In parallel, we extract directly involved semantic units to cover coarse, multi-entity events that
basic keyword matching may miss:

\begin{equation}
\label{eq:18}
\begin{aligned}
(\mathcal{G}_{\text{low}},\,\mathcal{C}_{\text{low}},\,\mathcal{S}_{\text{low}})
&= \mathrm{Retriever}_{\text{low}}\!\bigl(\mathcal{K}_{\text{low}};\\
&\quad \mathcal{G},\,\mathcal{C},\,\mathcal{S}\bigr),
\end{aligned}
\end{equation}

\begin{equation}
\label{eq:19}
\begin{aligned}
(\mathcal{G}_{\text{high}},\,\mathcal{C}_{\text{high}},\,\mathcal{S}_{\text{high}})
&= \mathrm{Retriever}_{\text{high}}\!\bigl(\mathcal{K}_{\text{high}};\\
&\quad \mathcal{G},\,\mathcal{C},\,\mathcal{S}\bigr),
\end{aligned}
\end{equation}

\noindent
where $\mathcal{K}_{\text{low}}$ / $\mathcal{K}_{\text{high}}$ are the low-/high-level keyword sets,
$\mathcal{G}$ is the knowledge graph, $\mathcal{C}$ the chunk set, and $\mathcal{S}$ the semantic-unit set. 
$\mathrm{Retriever}_{\text{low}}$ returns a keyword-matched, finely scoped subgraph $\mathcal{G}_{\text{low}}$
(with associated chunks $\mathcal{C}_{\text{low}}$ and units $\mathcal{S}_{\text{low}}$), 
while $\mathrm{Retriever}_{\text{high}}$ returns a theme-oriented subgraph $\mathcal{G}_{\text{high}}$
(with $\mathcal{C}_{\text{high}}$, $\mathcal{S}_{\text{high}}$), optionally expanded by lightweight graph heuristics
(e.g., short-hop neighbors or similarity-ranked additions) to include weak but informative context.

\indent When the candidates are insufficient, we augment them via similarity matching with semantic-level keywords:
\begin{equation}
\label{eq:20}
\mathcal{S}_{\text{sem}}
= \mathrm{Retriever}_{\text{sem}}\!\bigl(\mathcal{K}_{\text{sem}};\,\mathcal{S}\bigr),
\end{equation}

\begin{equation}
\label{eq:21}
\begin{aligned}
\mathcal{S}_{\text{all}}
&= \mathrm{Trim}\!\bigl(
   \mathcal{S}_{\text{low}} \cup \mathcal{S}_{\text{high}} \cup \mathcal{S}_{\text{sem}}
\bigr).
\end{aligned}
\end{equation}

\begin{equation}
\label{eq:22}
\begin{aligned}
\mathrm{Len}\!\bigl(\mathcal{S}_{\text{all}}\bigr)
&\le \max\!\Bigl(\tau,\;
\mathrm{Len}\!\bigl(\mathcal{S}_{\text{low}}\cup \mathcal{S}_{\text{high}}\bigr)\Bigr),
\end{aligned}
\end{equation}

\noindent
where $\mathcal{K}_{\text{sem}}$ is the semantic-level keyword set, 
$\mathrm{Retriever}_{\text{sem}}$ returns similarity-matched semantic units from $\mathcal{S}$,
$\mathrm{Trim}(\cdot)$ limits the set size, 
$\mathrm{Len}(\cdot)$ returns the set cardinality, and $\tau$ is a size threshold.

\indent Next, to further enrich both fine-grained binary relations and coarse-grained $n$-ary events, we traverse each semantic unit’s associated entities and relations:

\begin{equation}
\label{eq:23}
\begin{aligned}
(\mathcal{G}_{\text{sem}},\,\mathcal{C}_{\text{sem}})
&= \mathrm{Retriever}_{\mathcal{S}_{\text{all}}}\!\bigl(
   \mathcal{S}_{\text{all}};\, \mathcal{G},\, \mathcal{C}
\bigr),
\end{aligned}
\end{equation}

\begin{equation}
\label{eq:24}
\begin{aligned}
\mathcal{G}_{\text{all}}
&= \mathrm{Trim}\!\bigl(
   \mathcal{G}_{\text{low}} \cup \mathcal{G}_{\text{high}} \cup \mathcal{G}_{\text{sem}}
\bigr),
\end{aligned}
\end{equation}

\begin{equation}
\label{eq:25}
\begin{aligned}
\mathcal{C}_{\text{all}}
&= \mathrm{Trim}\!\bigl(
   \mathcal{C}_{\text{low}} \cup \mathcal{C}_{\text{high}} \cup \mathcal{C}_{\text{sem}}
\bigr),
\end{aligned}
\end{equation}

\noindent
where $\mathcal{S}_{\text{all}}$ is the aggregated semantic-unit set (from Eq.~\eqref{eq:21}); 
$\mathrm{Retriever}_{\mathcal{S}_{\text{all}}}(\cdot)$ collects a subgraph $\mathcal{G}_{\text{sem}}$ and chunk set $\mathcal{C}_{\text{sem}}$ by following entity/relation links and context indices associated with units in $\mathcal{S}_{\text{all}}$; 
$\mathrm{Trim}(\cdot)$ limits the size of the returned graph/chunk sets to a preset budget; 
$\mathcal{G}_{\text{low}},\mathcal{G}_{\text{high}},\mathcal{C}_{\text{low}},\mathcal{C}_{\text{high}}$ are from Eqs.~\eqref{eq:18}–\eqref{eq:19}.

\paragraph{Fusion for Generation}
Finally, we fuse retrieved snippets, semantic units, and graph context to guide the generator, producing an answer $R$ that cites fine-grained facts while maintaining global coherence across multi-entity events.

\begin{equation}
\label{eq:26}
\begin{aligned}
R &= \mathrm{AnsGen}_{\text{LLM}}\!\bigl(
      \mathcal{S}_{\text{all}},\,
      \mathcal{G}_{\text{all}},\,
      \mathcal{C}_{\text{all}}
    \bigr),
\end{aligned}
\end{equation}

\noindent
where $\mathcal{S}_{\text{all}}$ is the aggregated semantic-unit set (Eq.~\eqref{eq:21}), 
$\mathcal{G}_{\text{all}}$ the aggregated subgraph (Eq.~\eqref{eq:24}), 
$\mathcal{C}_{\text{all}}$ the aggregated chunk set (Eq.~\eqref{eq:25}); 
$\mathrm{AnsGen}_{\text{LLM}}(\cdot)$ denotes the LLM-based response generator conditioned on these inputs.

\section{Experiments}

\begin{table*}[t]
\centering
\small
\caption{Win rates (\%) of GOSU v.s. baselines across five datasets and four evaluation dimensions.}
\label{tab:1}
\begin{tabularx}{\textwidth}{@{} l *{12}{Y} @{}}
\toprule
 & \multicolumn{2}{c}{\textbf{Agriculture}} & \multicolumn{2}{c}{\textbf{CS}} & \multicolumn{2}{c}{\textbf{Hypertension}} & \multicolumn{2}{c}{\textbf{Legal}} & \multicolumn{2}{c}{\textbf{Mix}} \\
\cmidrule(lr){2-3} \cmidrule(lr){4-5} \cmidrule(lr){6-7} \cmidrule(lr){8-9} \cmidrule(lr){10-11}
 & {\scriptsize{NaiveRAG}} & {\scriptsize\textbf{GOSU}} & {\scriptsize{NaiveRAG}} & {\scriptsize\textbf{GOSU}} & {\scriptsize{NaiveRAG}} & {\scriptsize\textbf{GOSU}} & {\scriptsize{NaiveRAG}} & {\scriptsize\textbf{GOSU}} & {\scriptsize{NaiveRAG}} & {\scriptsize\textbf{GOSU}} & \scriptsize{Avg gap}\\
\midrule
\scriptsize\textbf Comprehensiveness & 12.0\% & \underline{\textbf{88.0\%}} & 14.3\% & \underline{\textbf{85.7\%}} & 11.1\% & \underline{\textbf{88.9\%}} & 12.0\% & \underline{\textbf{88.0\%}} & 23.1\% & \underline{\textbf{76.9\%}} & \textbf{+71.0\%}\\
\scriptsize\textbf Diversity         & 26.8\% & \underline{\textbf{73.2\%}} & 29.7\% & \underline{\textbf{70.3\%}} & 24.4\% & \underline{\textbf{75.6\%}} & 16.7\% & \underline{\textbf{83.3\%}} & 22.9\% & \underline{\textbf{77.1\%}} & \textbf{+51.8\%}\\
\scriptsize\textbf Empowerment       & 11.5\% & \underline{\textbf{88.5\%}} & 12.0\% & \underline{\textbf{88.0\%}} & 18.2\% & \underline{\textbf{81.8\%}} & 15.9\% & \underline{\textbf{84.1\%}} & 20.0\% & \underline{\textbf{80.0\%}} & \textbf{+69.0\%}\\
\scriptsize\textbf Overall           & 10.0\% & \underline{\textbf{90.0\%}} & 12.5\% & \underline{\textbf{87.5\%}} & 11.2\% & \underline{\textbf{88.5\%}} & 10.0\% & \underline{\textbf{90.0\%}} & 22.7\% & \underline{\textbf{77.3\%}} & \textbf{+73.4\%}\\

\cmidrule(lr){2-3} \cmidrule(lr){4-5} \cmidrule(lr){6-7} \cmidrule(lr){8-9} \cmidrule(lr){10-11}
 & {\scriptsize{LightRAG}} & {\scriptsize\textbf{GOSU}} & {\scriptsize{LightRAG}} & {\scriptsize\textbf{GOSU}} & {\scriptsize{LightRAG}} & {\scriptsize\textbf{GOSU}} & {\scriptsize{LightRAG}} & {\scriptsize\textbf{GOSU}} & {\scriptsize{LightRAG}} & {\scriptsize\textbf{GOSU}} & \scriptsize{Avg gap}\\

\midrule
\scriptsize\textbf Comprehensiveness & 8.8\% & \underline{\textbf{91.2\%}} & 20.0\% & \underline{\textbf{80.0\%}} & 25.0\% & \underline{\textbf{75.0\%}} & 29.4\% & \underline{\textbf{70.6\%}} & 41.7\% & \underline{\textbf{58.3\%}} & \textbf{+50.0\%}\\
\scriptsize\textbf Diversity         & 23.6\% & \underline{\textbf{76.4\%}} & 43.5\% & \underline{\textbf{56.5\%}} & 39.0\% & \underline{\textbf{61.0\%}} & 47.9\% & \underline{\textbf{52.1\%}}& 48.8\% & \underline{\textbf{51.2\%}} & \textbf{+18.9\%}\\
\scriptsize\textbf Empowerment       & 6.8\% & \underline{\textbf{93.2\%}} & 23.1\% & \underline{\textbf{76.9\%}} & 29.5\% & \underline{\textbf{70.5\%}} & 35.3\% & \underline{\textbf{64.7\%}} & 29.6\% & \underline{\textbf{70.4\%}} & \textbf{+50.3\%}\\
\scriptsize\textbf Overall           & 5.8\% & \underline{\textbf{94.2\%}} & 20.8\% & \underline{\textbf{79.2\%}} & 22.2\% & \underline{\textbf{77.8\%}} & 34.5\% & \underline{\textbf{65.5\%}} & 31.2\% & \underline{\textbf{68.8\%}} & \textbf{+54.2\%}\\

\cmidrule(lr){2-3} \cmidrule(lr){4-5} \cmidrule(lr){6-7} \cmidrule(lr){8-9} \cmidrule(lr){10-11}
 & {\scriptsize{HiRAG}} & {\scriptsize\textbf{GOSU}} & {\scriptsize{HiRAG}} & {\scriptsize\textbf{GOSU}} & {\scriptsize{HiRAG}} & {\scriptsize\textbf{GOSU}} & {\scriptsize{HiRAG}} & {\scriptsize\textbf{GOSU}} & {\scriptsize{HiRAG}} & {\scriptsize\textbf{GOSU}} & \scriptsize{Avg gap}\\

\midrule
\scriptsize\textbf Comprehensiveness & 14.3\% & \underline{\textbf{85.7\%}} & 42.9\% & \underline{\textbf{57.1\%}} & 20.0\% & \underline{\textbf{80.0\%}} & 33.3\% & \underline{\textbf{66.7\%}} & 33.3\% & \underline{\textbf{66.7\%}} & \textbf{+42.5\%}\\
\scriptsize\textbf Diversity         & 36.8\% & \underline{\textbf{63.2\%}} & 46.1\% & \underline{\textbf{53.9\%}} & 47.3\% & \underline{\textbf{52.7\%}} & 48.9\% & \underline{\textbf{51.1\%}} & 48.2\% & \underline{\textbf{51.8\%}} & \textbf{+9.1\%}\\
\scriptsize\textbf Empowerment       & 32.0\% & \underline{\textbf{68.0\%}} & 39.3\% & \underline{\textbf{60.7\%}} & 34.1\% & \underline{\textbf{65.9\%}} & 43.3\% & \underline{\textbf{56.7\%}} & 48.0\% & \underline{\textbf{52.0\%}} & \textbf{+21.3\%}\\
\scriptsize\textbf Overall           & 23.8\% & \underline{\textbf{76.2\%}} & 39.1\% & \underline{\textbf{60.9\%}} & 29.0\% & \underline{\textbf{71.0\%}} & 41.3\% & \underline{\textbf{58.7\%}} & 45.5\% & \underline{\textbf{54.5\%}} & \textbf{+28.5\%} \\

\cmidrule(lr){2-3} \cmidrule(lr){4-5} \cmidrule(lr){6-7} \cmidrule(lr){8-9} \cmidrule(lr){10-11}
 & \multicolumn{1}{>{\raggedright\arraybackslash}Y}{\tiny HyperGraphRAG}
 & \multicolumn{1}{>{\raggedleft\arraybackslash}Y}{\tiny\textbf{GOSU}}
 & \multicolumn{1}{>{\raggedright\arraybackslash}Y}{\tiny HyperGraphRAG}
 & \multicolumn{1}{>{\raggedleft\arraybackslash}Y}{\tiny\textbf{GOSU}}
 & \multicolumn{1}{>{\raggedright\arraybackslash}Y}{\tiny HyperGraphRAG}
 & \multicolumn{1}{>{\raggedleft\arraybackslash}Y}{\tiny\textbf{GOSU}}
 & \multicolumn{1}{>{\raggedright\arraybackslash}Y}{\tiny HyperGraphRAG}
 & \multicolumn{1}{>{\raggedleft\arraybackslash}Y}{\tiny\textbf{GOSU}}
 & \multicolumn{1}{>{\raggedright\arraybackslash}Y}{\tiny HyperGraphRAG}
 & \multicolumn{1}{>{\raggedleft\arraybackslash}Y}{\tiny\textbf{GOSU}} & \scriptsize{Avg gap}\\

\midrule
\scriptsize\textbf Comprehensiveness & 5.9\% & \underline{\textbf{94.1\%}} & 3.7\% & \underline{\textbf{96.3\%}} & 6.5\% & \underline{\textbf{93.5\%}} & 5.6\% & \underline{\textbf{94.4\%}} & 7.7\% & \underline{\textbf{92.3\%}} & \textbf{+88.2\%}\\
\scriptsize\textbf Diversity         & 26.2\% & \underline{\textbf{73.8\%}} & 14.6\% & \underline{\textbf{85.4\%}} & 29.8\% & \underline{\textbf{70.2\%}} & 17.9\% & \underline{\textbf{82.1\%}} & 21.5\% & \underline{\textbf{78.5\%}} & \textbf{+56.0\%}\\
\scriptsize\textbf Empowerment       & 7.9\% & \underline{\textbf{92.1\%}} & 12.0\% & \underline{\textbf{88.0\%}} & 27.5\% & \underline{\textbf{72.5\%}} & 21.4\% & \underline{\textbf{78.6\%}} & 5.0\% & \underline{\textbf{95.0\%}} & \textbf{\textbf{+70.5\%}}\\
\scriptsize\textbf Overall           & 3.4\% & \underline{\textbf{96.6\%}} & 12.5\% & \underline{\textbf{87.5\%}} & 20.8\% & \underline{\textbf{79.2\%}} & 7.9\% & \underline{\textbf{83.3\%}} & 5.6\% & \underline{\textbf{94.4\%}} & \textbf{+78.2\%}\\
\bottomrule
\end{tabularx}
\end{table*}

\indent To comprehensively assess the effectiveness of GOSU on knowledge-intensive generation tasks, we conducted extensive experiments on several publicly available domain datasets, compared GOSU against a range of representative baselines, and performed systematic ablation studies.

\subsection{Experimental Setup}

\paragraph{Datasets.}To evaluate GOSU’s cross-vertical performance and follow established experimental protocols~\citep{guo2024lightrag, luo2025hypergraphrag}, we selected four domain datasets from the UltraDomain benchmark~\citep{qian2024memorag}: Agriculture, Computer Science (CS), Law, and Mix, together with a fifth dataset consisting of the most recent international hypertension guidelines~\citep{mccarthy2025escpharm}. Additionally, following the generation methodology of Edge et al.~\citep{edge2024graphragqfs}, we employed an LLM to synthesize distinct RAG user profiles for each vertical and, from each profile’s perspective, generated multiple corpus-level queries that require holistic comprehension of the entire collection.

\paragraph{Baselines.}We compared GOSU against four state-of-the-art public RAG systems: NaiveRAG~\citep{gao2023ragsurvey}, the standard baseline that retrieves fixed-length text chunks by similarity; LightRAG~\citep{guo2024lightrag}, a lightweight model that employs a two-tier retrieval strategy to balance recall and efficiency; HiRAG~\citep{huang2025hierarchicalrag}, a framework that leverages hierarchical knowledge representations to enhance semantic understanding and capture structural relations; and HyperGraphRAG~\citep{luo2025hypergraphrag}, a novel RAG approach that incorporates hypergraph structures to capture higher-order, multi-entity relations.

\paragraph{Evaluation Metrics.}To more thoroughly evaluate outcomes, particularly for queries that invoke complex, high-level semantics, we follow the evaluation protocol of KnowTuning et al.~\citep{lyu2024knowtuning,guo2024lightrag,edge2024graphragqfs} and adopt four assessment dimensions: Comprehensiveness, Diversity, Empowerment, and Overall. To ensure evaluation accuracy and mitigate potential positional bias~\citep{zheng2023llmasajudge,pezeshkpour2023optionorder}, we employed an alternating pairwise comparison protocol in which candidate answers were presented in randomized left–right order and judged pairwise; for each evaluation dimension we selected the preferred answer based on these pairwise judgments. Specifically, we accept a comparison outcome only when the alternating pairwise judgments produce a consistent preference; if they do not, we treat the observed quality difference as being below the level of positional bias, deem the result inconclusive, and exclude it from further analysis.
 The final overall preference was determined by aggregating the rankings across the three primary dimensions (Comprehensiveness, Diversity, and Insightfulness), with ties resolved using the Overall Quality score.

\paragraph{Implementation Details.}We used GPT-4o-mini as the generative model and BGE-m3 for vector embeddings. To ensure experimental consistency and fair comparison, chunk size and all other retrieval- and generation-related hyperparameters were held identical across all methods.

\begin{figure}[t]
  \centering
  \includegraphics[width=\columnwidth]{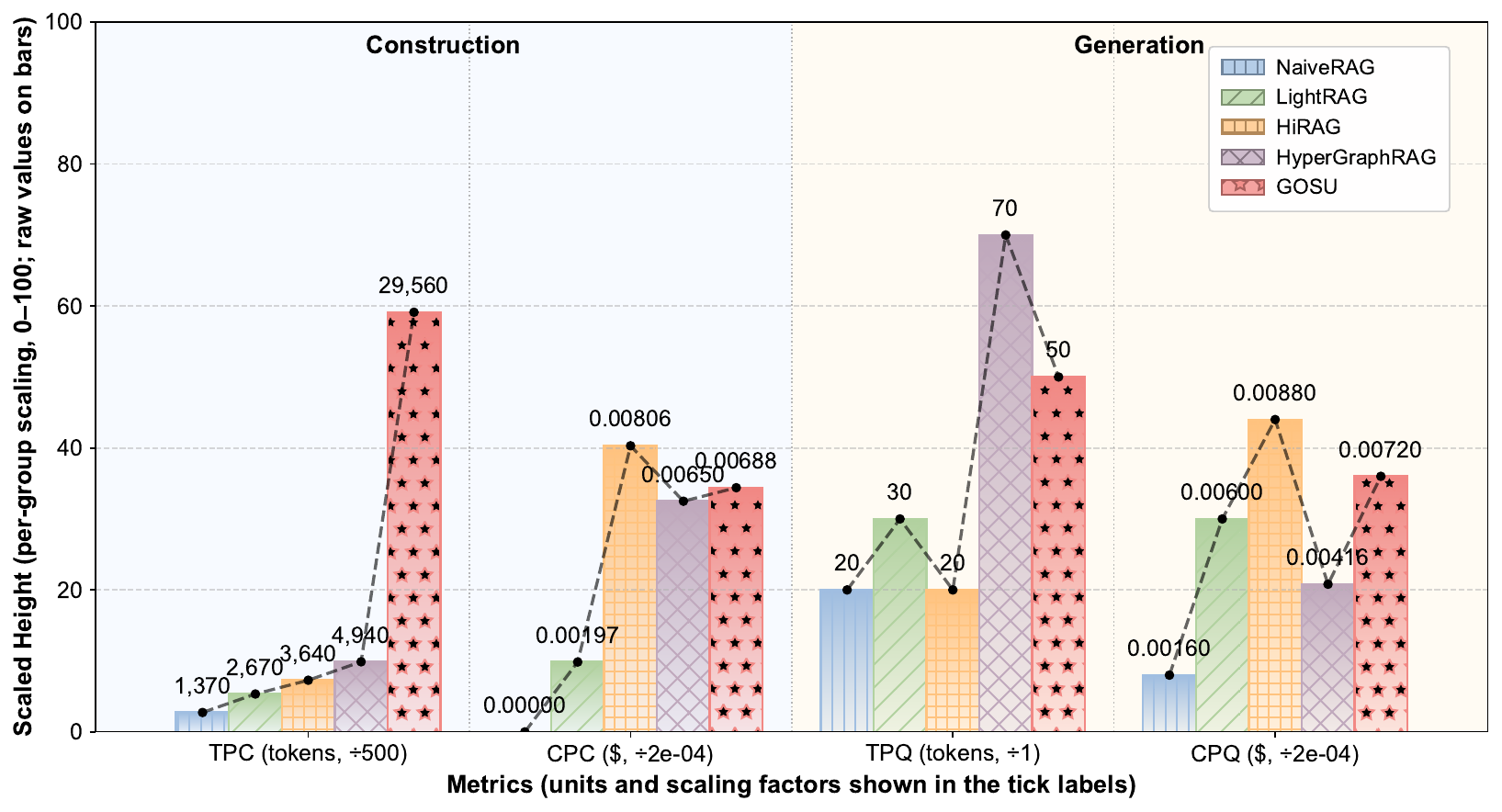}
  \caption{Tokens comparision and cost comparison of GOSU and baselines for Knowledge Construction versus Retrieval and Generation. TPB/TPQ: tokens per chunk/query (tokens). CPB/CPQ: cost per chunk/query (\$).}
  \label{fig:3}
\end{figure}

\begin{table*}[t]
\centering
\small
\caption{Win rates (\%) of GOSU vs. its ablated versions across five datasets and four evaluation dimensions.}
\label{tab:2}
\begin{tabularx}{\textwidth}{@{} l *{12}{Y} @{}}

\toprule
& \multicolumn{10}{c}{\textbf{Construction}}\\
\midrule

 & \multicolumn{2}{c}{\textbf{Agriculture}} & \multicolumn{2}{c}{\textbf{CS}} & \multicolumn{2}{c}{\textbf{Hypertension}} & \multicolumn{2}{c}{\textbf{Legal}} & \multicolumn{2}{c}{\textbf{Mix}} \\

\cmidrule(lr){2-3} \cmidrule(lr){4-5} \cmidrule(lr){6-7} \cmidrule(lr){8-9} \cmidrule(lr){10-11}
  & {\scriptsize{w/o~GO}} & {\scriptsize\textbf{GOSU}} & {\scriptsize{w/o~GO}} & {\scriptsize\textbf{GOSU}} & {\scriptsize{w/o~GO}} & {\scriptsize\textbf{GOSU}} & {\scriptsize{w/o~GO}} & {\scriptsize\textbf{GOSU}} & {\scriptsize{w/o~GO}} & {\scriptsize\textbf{GOSU}} & \scriptsize{Avg~gap}\\
\midrule
\scriptsize\textbf Comprehensiveness & 42.9\% & \underline{\textbf{57.1\%}} & 41.7\% & \underline{\textbf{58.3\%}} & 25.0\% & \underline{\textbf{75.0\%}} & 14.3\% & \underline{\textbf{85.7\%}} & 20.0\% & \underline{\textbf{80.0\%}} & \textbf{+42.4\%}\\
\scriptsize\textbf Diversity         & 46.7\% & \underline{\textbf{53.3\%}} & 48.7\% & \underline{\textbf{51.3\%}} & 48.1\% & \underline{\textbf{51.9\%}} & 40.3\% & \underline{\textbf{59.7\%}} & 49.4\% & \underline{\textbf{50.6\%}} & \textbf{+6.7\%}\\
\scriptsize\textbf Empowerment       & 46.4\% & \underline{\textbf{53.6\%}} & 42.1\% & \underline{\textbf{57.9\%}} & 42.9\% & \underline{\textbf{57.1\%}} & 22.7\% & \underline{\textbf{77.3\%}} & 40.9\% & \underline{\textbf{59.1\%}} & \textbf{\textbf{+22.0\%}}\\
\scriptsize\textbf Overall           & 41.7\% & \underline{\textbf{58.3\%}} & 41.2\% & \underline{\textbf{58.8\%}} & 32.4\% & \underline{\textbf{67.6\%}} & 15.8\% & \underline{\textbf{84.2\%}} & 37.5\% & \underline{\textbf{62.5\%}} & \textbf{+32.6\%}\\

\midrule
 & \multicolumn{10}{c}{\textbf{Retrieval~\&~Generation}}\\
\midrule

 & \multicolumn{2}{c}{\textbf{Agriculture}} & \multicolumn{2}{c}{\textbf{CS}} & \multicolumn{2}{c}{\textbf{Hypertension}} & \multicolumn{2}{c}{\textbf{Legal}} & \multicolumn{2}{c}{\textbf{Mix}} \\
\cmidrule(lr){2-3} \cmidrule(lr){4-5} \cmidrule(lr){6-7} \cmidrule(lr){8-9} \cmidrule(lr){10-11}

 & {\scriptsize{w/o~EL}} & {\scriptsize\textbf{GOSU}} & {\scriptsize{w/o~EL}} & {\scriptsize\textbf{GOSU}} & {\scriptsize{w/o~EL}} & {\scriptsize\textbf{GOSU}} & {\scriptsize{w/o~EL}} & {\scriptsize\textbf{GOSU}} & {\scriptsize{w/o~EL}} & {\scriptsize\textbf{GOSU}} & \scriptsize{Avg~gap}\\
\midrule

\scriptsize\textbf Comprehensiveness & 46.7\% & \underline{\textbf{53.3\%}} & 38.5\% & \underline{\textbf{61.5\%}} & 47.1\% & \underline{\textbf{52.9\%}} & 37.5\% & \underline{\textbf{62.5\%}} & 46.2\% & \underline{\textbf{53.8\%}} & \textbf{+13.6\%}\\
\scriptsize\textbf Diversity         & 45.4\% & \underline{\textbf{54.6\%}} & 41.6\% & \underline{\textbf{58.4\%}} & 48.3\% & \underline{\textbf{51.7\%}} & 45.5\% & \underline{\textbf{54.5\%}} & 43.5\% & \underline{\textbf{56.5\%}} & \textbf{+10.3\%}\\
\scriptsize\textbf Empowerment       & 44.4\% & \underline{\textbf{55.6\%}} & 36.0\% & \underline{\textbf{64.0\%}} & 44.6\% & \underline{\textbf{55.4\%}} & 47.4\% & \underline{\textbf{52.6\%}} & 43.5\% & \underline{\textbf{56.5\%}} & \textbf{+13.6\%}\\
\scriptsize\textbf Overall           & 45.2\% & \underline{\textbf{54.8\%}} & 42.9\% & \underline{\textbf{57.1\%}} & 46.9\% & \underline{\textbf{53.1\%}} & 47.1\% & \underline{\textbf{52.9\%}} & 45.0\% & \underline{\textbf{55.0\%}} & \textbf{+9.2\%}\\
 \cmidrule(lr){2-3} \cmidrule(lr){4-5} \cmidrule(lr){6-7} \cmidrule(lr){8-9} \cmidrule(lr){10-11}

 & {\scriptsize{w/o~RL}} & {\scriptsize\textbf{GOSU}} & {\scriptsize{w/o~RL}} & {\scriptsize\textbf{GOSU}} & {\scriptsize{w/o~RL}} & {\scriptsize\textbf{GOSU}} & {\scriptsize{w/o~RL}} & {\scriptsize\textbf{GOSU}} & {\scriptsize{w/o~RL}} & {\scriptsize\textbf{GOSU}} & \scriptsize{Avg~gap}\\

\midrule

\scriptsize\textbf Comprehensiveness & 45.5\% & \underline{\textbf{54.5\%}} & 44.4\% & \underline{\textbf{55.6\%}} & 33.3\% & \underline{\textbf{66.7\%}} & 36.4\% & \underline{\textbf{63.6\%}} & 47.1\% & \underline{\textbf{52.9\%}} & \textbf{+17.3\%}\\
\scriptsize\textbf Diversity         & 44.9\% & \underline{\textbf{55.1\%}} & 46.3\% & \underline{\textbf{53.7\%}} & 40.5\% & \underline{\textbf{59.5\%}} & 45.8\% & \underline{\textbf{54.2\%}}& 39.2\% & \underline{\textbf{60.8\%}} & \textbf{+13.3\%}\\
\scriptsize\textbf Empowerment       & 45.7\% & \underline{\textbf{54.3\%}} & 40.0\% & \underline{\textbf{60.0\%}} & 37.3\% & \underline{\textbf{62.7\%}} & 48.0\% & \underline{\textbf{52.0\%}} & 41.2\% & \underline{\textbf{58.8\%}} & \textbf{+15.1\%}\\
\scriptsize\textbf Overall           & 45.8\% & \underline{\textbf{54.2\%}} & 46.7\% & \underline{\textbf{53.3\%}} & 35.1\% & \underline{\textbf{64.9\%}} & 45.0\% & \underline{\textbf{55.0\%}} & 40.9\% & \underline{\textbf{59.1\%}} & \textbf{+14.6\%}\\

\cmidrule(lr){2-3} \cmidrule(lr){4-5} \cmidrule(lr){6-7} \cmidrule(lr){8-9} \cmidrule(lr){10-11}

& {\tiny{w/o~EL~\&~RL}} & {\scriptsize\textbf{GOSU}} & {\tiny{w/o~EL~\&~RL}} & {\scriptsize\textbf{GOSU}} & {\tiny{w/o~EL~\&~RL}} & {\scriptsize\textbf{GOSU}} & {\tiny{w/o~EL~\&~RL}} & {\scriptsize\textbf{GOSU}} & {\tiny{w/o~EL~\&~RL}} & {\scriptsize\textbf{GOSU}} & \scriptsize{Avg~gap}\\

\midrule

\scriptsize\textbf Comprehensiveness & 33.3\% & \underline{\textbf{66.7\%}} & 35.7\% & \underline{\textbf{64.3\%}} & 31.2\% & \underline{\textbf{68.8\%}} & 43.7\% & \underline{\textbf{56.3\%}} & 36.8\% & \underline{\textbf{63.2\%}} & \textbf{+27.7\%}\\
\scriptsize\textbf Diversity         & 32.3\% & \underline{\textbf{67.7\%}} & 38.7\% & \underline{\textbf{61.3\%}} & 31.7\% & \underline{\textbf{68.3\%}} &46.3\% & \underline{\textbf{53.7\%}} & 34.7\% & \underline{\textbf{65.3\%}} & \textbf{+26.5\%}\\
\scriptsize\textbf Empowerment       & 30.6\% & \underline{\textbf{69.4\%}} & 37.5\% & \underline{\textbf{62.5\%}} & 43.1\% & \underline{\textbf{56.9\%}} & 48.1\% & \underline{\textbf{51.9\%}} & 37.9\% & \underline{\textbf{62.1\%}} & \textbf{+21.1\%}\\
\scriptsize\textbf Overall           & 35.7\% & \underline{\textbf{64.3\%}} & 27.8\% & \underline{\textbf{72.2\%}} & 42.4\% & \underline{\textbf{57.6\%}} & 45.8\% & \underline{\textbf{54.2\%}} & 33.3\% & \underline{\textbf{66.7\%}} & \textbf{+26.0\%} \\

\cmidrule(lr){2-3} \cmidrule(lr){4-5} \cmidrule(lr){6-7} \cmidrule(lr){8-9} \cmidrule(lr){10-11}

 & {\scriptsize{w/o~SL}} & {\scriptsize\textbf{GOSU}} & {\scriptsize{w/o~SL}} & {\scriptsize\textbf{GOSU}} & {\scriptsize{w/o~SL}} & {\scriptsize\textbf{GOSU}} & {\scriptsize{w/o~SL}} & {\scriptsize\textbf{GOSU}} & {\scriptsize{w/o~SL}} & {\scriptsize\textbf{GOSU}} & \scriptsize{Avg~gap}\\

\midrule

\scriptsize\textbf Comprehensiveness & 45.5\% & \underline{\textbf{54.5\%}} & 22.2\% & \underline{\textbf{77.8\%}} & 35.3\% & \underline{\textbf{64.7\%}} & 42.9\% & \underline{\textbf{57.1\%}} & 18.2\% & \underline{\textbf{81.8\%}} & \textbf{+34.4\%}\\
\scriptsize\textbf Diversity         & 38.3\% & \underline{\textbf{61.7\%}} & 43.4\% & \underline{\textbf{56.6\%}} & 44.8\% & \underline{\textbf{55.2\%}} &42.2\% & \underline{\textbf{57.8\%}} & 39.2\% & \underline{\textbf{60.8\%}} & \textbf{+16.8\%}\\
\scriptsize\textbf Empowerment       & 38.1\% & \underline{\textbf{61.9\%}} & 32.0\% & \underline{\textbf{68.0\%}} & 38.5\% & \underline{\textbf{61.5\%}} & 34.8\% & \underline{\textbf{65.2\%}} & 36.0\% & \underline{\textbf{64.0\%}} & \textbf{+28.2\%}\\
\scriptsize\textbf Overall           & 38.9\% & \underline{\textbf{61.1\%}} & 25.0\% & \underline{\textbf{75.0\%}} & 39.4\% & \underline{\textbf{60.6\%}} & 36.8\% & \underline{\textbf{63.2\%}} & 38.1\% & \underline{\textbf{61.9\%}} & \textbf{+28.7\%} \\

\bottomrule
\end{tabularx}
\end{table*}

\subsection{Experimental Results}

\indent We compared GOSU against the baseline methods across each domain along multiple evaluation dimensions; the results are summarized in Table~\ref{tab:1} and Table~\ref{tab:2}.

\paragraph{General Comparison.}As shown in Table~\ref{tab:1}, GOSU demonstrated stable performance across domains, consistently achieving higher win rates than all baselines on the four evaluation dimensions—Comprehensiveness, Diversity, Empowerment, and Overall—indicating its superior ability to produce more complete, varied, and practically useful responses. 

\indent Compared to NaiveRAG, GOSU achieved an average win-rate margin exceeding 50\% across all evaluation dimensions, highlighting the superiority of graph-based RAG approaches over chunk-based retrieval in capturing complex semantic dependencies for knowledge-intensive tasks. Although GOSU is also a graph-augmented RAG method, it consistently outperforms LightRAG and HyperGraphRAG. This result indicates that, compared with approaches that rely solely on pairwise edges or purely $n$-ary hyperedges, GOSU more effectively integrates fine-grained and coarse-grained semantic units and leverages them during retrieval and generation to produce higher-quality responses. Among the baselines, HiRAG exhibits the smallest performance gap relative to GOSU. This finding indicates that hierarchical knowledge representations do enhance semantic understanding and structural capture, and it further validates that GOSU’s strategy—driving the pipeline with globally completed semantic units—can achieve comparable or even superior results by explicitly integrating corpus-level semantic coherence with a dual-phase retrieval-and-generation process. 

\indent Experiments across multiple datasets highlight GOSU’s superior capability to integrate semantic information and to recognize structural variations across tasks and domains, with particularly strong gains in knowledge-intensive scenarios.

\paragraph{Ablation Study.}To rigorously assess the contribution of each component within the GOSU framework, we conducted extensive ablation studies at both the knowledge construction stage and the retrieval and generation stage (see Table~\ref{tab:2}). 

\indent Knowledge Construction Stage. We ablated the global-level semantic unit optimization (w/o GO). This modification produced pronounced declines in Comprehensiveness, Empowerment, and Overall scores, with the performance degradation particularly marked on the medical and legal benchmarks. These results indicate that the GO module is crucial for globally extracting and completing $n$-ary relations and for facilitating the identification of relevant binary relations, thereby improving evidence aggregation and downstream generation quality.

\indent Retrieval and Generation stage. We ablated each component of the three-stage retrieval mechanism—removing the entity layer (w/o EL), the relation layer (w/o RL), and the semantic-unit layer (w/o SL). All ablations produced measurable performance degradations, with the removal of the semantic-unit layer yielding the largest decline. This result validates the critical role of semantic units in completing coarse-grained information and supporting robust multi-entity evidence aggregation. Additionally, we ablated both the entity and relation layers (w/o EL \& RL). The combined removal produced a marked performance degradation, further confirming that fine-grained knowledge—encoded by entity- and relation-level signals—is also indispensable for producing high-quality, factually grounded generations.

\indent These experimental results demonstrate that each module deployed across GOSU’s pipeline is necessary to achieve optimal generation quality.

\paragraph{Analysis of Efficency and Cost.}We conducted a comprehensive cost comparison of GOSU and four baseline methods across the knowledge construction (offline) and retrieval and generation (online) phases. The experimental results are presented in~Fig.\ref{fig:3}. We measured four cost metrics: token consumption for vector embeddings per text chunk (TPC), prompt-completion cost per text chunk (CPC), token consumption for vector embeddings per query (TPQ), and prompt-completion cost per query (CPQ).

\indent During the knowledge construction phase, the integration of the global semantic unit optimization module increased GOSU’s embedding token consumption: total embedding token consumption (TPC) reached 29,560 tokens, substantially higher than HyperGraphRAG’s 4,940 tokens. In terms of CPC (cost per completion), GOSU incurred \$0.00688 per completion—marginally higher than the three other baselines but still lower than HiRAG, which incurred \$0.00806. During the generation phase, GOSU recorded a TPQ of 50 tokens, slightly higher than LightRAG (30 tokens) but lower than HyperGraphRAG (70 tokens). Additionally, for CPQ (cost per completion at query time), GOSU incurred \$0.00720 per completion—comparable to LightRAG (\$0.00600) and lower than HiRAG (\$0.00880). These results indicate that, although GOSU incurs additional cost to better accommodate knowledge-intensive tasks, the overhead remains within acceptable bounds. By employing a coarse-to-fine, two-tier filtering strategy to globally complete semantic units, GOSU balances structural and semantic representations, uncovers both fine-grained binary relations and coarse-grained $n$-ary relations, and thereby effectively supports a three-stage retrieval-and-generation pipeline. Moreover, because the usage costs of many high-performance models, including those employed in this study, have been steadily decreasing and in some cases become free, the additional token overhead is acceptable.

\indent Regarding efficiency, because GOSU performs pairwise similarity comparisons between semantic units during knowledge-graph construction, it incurs substantial computational overhead and leads to increased preprocessing time—especially on large corpora. This pairwise matching step is computationally intensive compared with simpler indexing strategies and represents the primary source of GOSU's higher offline latency. Nonetheless, knowledge-graph construction is typically a one-time operation in most deployment scenarios, and subsequent system activity concentrates on retrieval and query handling; therefore, the upfront cost does not materially degrade ongoing online efficiency.

\section{Conclusion}

\indent GOSU is a new framework that centers the RAG pipeline on semantic units which are optimized at the corpus level; by using these globally consistent units to guide the extraction of binary and $n$-ary relations, GOSU achieves a balanced fusion of fine-grained entity links and coarse-grained multi-entity structures, resulting in more faithful, comprehensive, and coherent retrieval-augmented generation. Unlike approaches limited to individual text chunks, GOSU introduces a two-stage coarse-to-fine filtering mechanism to better summarize semantic units and extract structural information at the global corpus level. Driven by these semantic units, the SU-centric design and three-stage retrieval pipeline supplement low-level signals with tightly related high-level perspectives, yielding more coherent and semantically complete retrieval and generation. Extensive experiments demonstrate that GOSU consistently outperforms existing RAG pipelines across diverse vertical domains and related tasks. Although GOSU incurs additional computational and monetary costs to achieve improved performance, these trade-offs remain within acceptable bounds. While GOSU sacrifices some efficiency and incurs higher preprocessing and token costs to deliver superior retrieval and generation quality, our measurements show that these overheads are moderate and justified bythe overall substantial gains. Taken together, GOSU emphasizes the equal and complementary roles of fine- and coarse-grained relational modeling within graph-based RAG frameworks, delivering a scalable and high-quality approach for real-world AIGC tasks that require faithful, comprehensive, and coherent knowledge integration.

\section*{Limitations}

Cross-domain experiments demonstrate that GOSU achieves substantial improvements in retrieval-augmented generation, but there remains room for further refinement.

\begin{itemize}[leftmargin=*, itemsep=1ex]
  \item First, the current method does not incorporate multimodal inputs and therefore cannot fully exploit knowledge embedded in images, tables, and other non-textual artifacts within multimodal corpora, which may lead to omission of important information.
  \item Second, GOSU primarily focuses on $n$-ary and binary relations and may lack the capacity to uncover deep chains of reasoning required for more complex inferential tasks.
  \item Additionally, although GOSU enriches knowledge structure by centering on semantic units and employing a three-layer retrieval pipeline, there is still potential to further improve retrieval and generation efficiency.
\end{itemize}

Future work will investigate methods to overcome the limitations identified above.

\section*{Ethics Statement}

\indent This paper investigates RAG via GOSU, a semantic-unit–centric framework that globally optimizes semantic units to drive extraction of binary and $n$-ary relations. We employ large language models for semantic-unit extraction and SU-centric graph construction, together with retrieval-augmented generation techniques to improve knowledge representation and generation quality. All data used in this study are publicly available and contain no personally identifiable or sensitive information; therefore, we believe the work adheres to ethical principles.

\bibliographystyle{acl_natbib}
\bibliography{custom} 

\end{document}